\documentclass[11pt,a4paper]{article}
\usepackage{authblk}
\usepackage[hyperref]{naaclhlt2019}
\usepackage{times}
\usepackage{latexsym}
\usepackage{todonotes}

\usepackage{url}

\usepackage{pgfplots,tikz}
\pgfplotsset{compat=newest}
\usetikzlibrary{intersections}
\usepackage{ wasysym }
\usetikzlibrary{decorations.markings}

\usepackage{todonotes}

\usepackage[skip=3.5pt]{caption}
\aclfinalcopy 
\def\aclpaperid{1638} 

\title{Crowdsourcing Lightweight Pyramids \\ 
for Manual Summary Evaluation}

\author[1]{\bf Ori Shapira}
\author[1]{\bf David Gabay}
\author[2]{\bf Yang Gao}
\author[1]{\bf Hadar Ronen}
\author[3]{\\ \bf Ramakanth Pasunuru}
\author[3]{\bf Mohit Bansal}
\author[1]{\bf Yael Amsterdamer}
\author[1]{\bf Ido Dagan}
\affil[1]{Bar-Ilan University}
\affil[2]{UKP Technische Universit{\" a}t Darmstadt}
\affil[3]{UNC Chapel Hill}
\affil[  ]{\tt \small \{obspp18, dawid.gabay\}@gmail.com, gao@ukp.informatik.tu-darmstadt.de}
\affil[  ]{\tt \small hadarg@gmail.com, \{ram, mbansal\}@cs.unc.edu, \{amstery, dagan\}@cs.biu.ac.il}

\date{}

\begin{document}
\maketitle
\begin{abstract}
Conducting a manual evaluation is considered an essential part of summary evaluation methodology. Traditionally, the Pyramid protocol, which exhaustively compares system summaries to references, has been perceived as very reliable, providing objective scores. Yet, due to the high cost of the Pyramid method and the required expertise, researchers resorted to cheaper and less thorough manual evaluation methods, such as Responsiveness and pairwise comparison, attainable via crowdsourcing. We revisit the Pyramid approach, proposing a lightweight sampling-based version that is crowdsourcable. We analyze the performance of our method in comparison to original expert-based Pyramid evaluations, showing higher correlation relative to the common Responsiveness method.
We release our crowdsourced Summary-Content-Units, along with all crowdsourcing scripts, for future evaluations.

\end{abstract}

\section{Introduction}
Evaluating \textit{content} quality of summaries is an integral part of summarization research. Measuring the performance of a summarization system can be done through either automatic or manual evaluation. An automatic evaluation, in practice working at the lexical level, provides an inexpensive means of measuring the validity of a system, both for system comparisons and for quick development cycle testing.
Due to the shallowness of the automatic approaches, their reliability is often perceived as insufficient \citep{owczarzak2012autoEvalAssessment, chaganty2018price}. This calls for the more expensive manual evaluation, which employs human-in-the-loop protocols for assessment.

The Pyramid method \citep{nenkova2004pyramids} is a prominent manual evaluation methodology that is considered highly reliable for comparing summarization systems. It relies on a small set of manually-crafted reference summaries, out of which all summary content units (SCUs) are manually extracted. System summaries are then manually checked for coverage of each individual SCU, from which an overall system score is derived.
The Pyramid evaluation method's reliability comes at a cost. It requires laborious manual work performed by annotators who must browse through non-trivial guidelines \citep{Passonneau2006PyramidGuide}. Due to these drawbacks, it was only used in a few DUC and TAC \citep{Nist2014Duc, Nist2018Tac} benchmarks.

Instead, summarization work in recent years has mostly employed simpler manual evaluation approaches, such as Responsiveness and pairwise comparison, which do not rely on reference summaries and can be attained via crowdsourcing.
Yet, these methods are quite subjective, since evaluators need to provide only a single global judgment for the quality of a summary (or a pair of summaries). Such judgments are far more subjective than the Pyramid score, which is derived from many, more objective, local decisions, each judging independently the presence of an individual SCU. Indeed, it was shown that the above subjective crowdsourcing-based evaluation methods are not reliable enough to produce consistent scores across experiments \citep{gillick2010crowdsourcing}.

We propose a simplified crowdsourcable and reproducible version of the Pyramid method, that suggests appealing advantages over prior crowdsourcable evaluation methods. Like the original Pyramid, our method leverages the strong signal of the reference summaries and similarly bases its score on less subjective SCU judgments.
In contrast to the original Pyramid, we rely on statistical sampling rather than exhaustive SCU extraction and testing, lowering overall cost. Empirically, our method correlates with the original Pyramid scores better than the common Responsiveness method, and shows better stability.

\section{Background: Manual Summary Evaluation}
The Pyramid method \citep{nenkova2004pyramids} consists of two manual phases. The first phase is \textit{pyramid creation}, performed once when a dataset is constructed, per each input topic to be summarized (either a single document or a set of documents). In this phase, experts exhaustively extract all \textit{SCU contributors} (``mentions"), each being a text span describing an individual fact.
SCU contributors are extracted from \emph{several} reference summaries of the source text.
Coreferring SCU contributors across reference summaries are then merged into a single \textit{SCU}, which is given a representative label. Each SCU is then assigned a weight, equal to the number of reference summaries in which it was found, indicating its salience. 

The second phase is \textit{system evaluation}, performed over the summaries produced by the evaluated system. Each Pyramid SCU for the source text is manually checked for its presence in the given system summary, whose Pyramid score is then computed as a normalized sum of the weights of the SCUs it contains. The overall system score is defined as the average Pyramid score over all its evaluated summaries.
Although certain normalization variants attempt to weigh in SCU precision, the score is essentially an absolute ``recall-style'' interpretation reflecting the system's ability to cover the content units found in the reference summaries. 
Such a fairly robust score allows, in principle, system comparison across experiments \citep{nenkova2004pyramids}.

We note that due to the Pyramid method's reliability, some research has been carried out on simulating the Pyramid method as a fully automatic one \citep{yang2016PEAK, hirao2018autoPyramids}. The hope of such a line of work is to find an automatic evaluation method that is more reliable than the commonly used ones, by taking the reference summary \textit{semantic} content into account.
Despite these efforts, automated Pyramid evaluations did not make their way yet to mainstream summary evaluation practices, where variants of the ROUGE metric \citep{lin2004rouge} still prevail. In any case, as this paper focuses on manual evaluation, we compare our results to those of the manual Pyramid.

The \emph{Responsiveness} method, introduced in DUC 2003 \citep{Nist2003Duc2003Tasks}, does not require reference summaries. Instead, human evaluators typically read both the source text and the system summary. They then assign a single subjective score on a Likert scale for the summary quality, often with respect to a topic statement or guiding question. Finally, compared systems are ranked by the average score of their summaries. This method naturally developed into a crowdsourcing task, and is now used frequently in some variants \citep{Grusky2018Newsroom, paulus2017deep}.

Another common crowdsourcable evaluation method is pairwise comparison \citep{Gao2018Preference,Falke2017ConceptMapMDS,Fan2018ControllableAS}:
an evaluator is asked to judge which of two competing summaries of the same text is superior, usually while observing the source text.
This protocol allows comparing only two systems at a time, where the superior is determined by the total votes over all input texts. The obvious disadvantage of the approach is the difficulty of comparing many systems, in the absence of absolute scores. Also, this method may tend to suffer from transitivity inconsistencies when comparing multiple system pairs \citep{gillick2010crowdsourcing}.

The lightweight crowdsourcable Pyramid version we propose aims to preserve the interpretability and relative objectiveness of the Pyramid scores. This could provide absolute scores for comparing multiple systems, which the pairwise method does not, in a more reliable manner than Responsiveness evaluation.
\label{sec_background}

\section{Our Lightweight Pyramid Method}
Our Lightweight Pyramid method mimics the two phases of the original Pyramid protocol in a crowdsourced setting, with some adjustments.

\paragraph*{Pyramid creation.}
The input for this phase is several reference summaries of a topic.
Each reference is presented to two crowd workers, asking to extract eight SCU-like statements, yielding 16 potential SCUs per reference summary. The instructions guide workers to copy-and-paste extractions from the text, possibly modifying them to stand-alone sentences, that should (a) be brief and focused on a single fact; (b) capture important information; (c) rely solely on the text rather than general knowledge of the worker.
Further, the statements should appear in different places in the text.

The copy-and-paste approach allows us to easily detect and filter duplicate statements extracted from the \textit{same} reference by both annotators, which we identify via bag-of-lemmas cosine similarity. Further, too long sentences are filtered. 
In our experiments (see Section \ref{sec_results}), we were left with an average of about 13 SCUs per reference summary. 
Then, we take the union of SCUs from all reference summaries, which yielded in our experiments 51 SCUs on average per topic, coming from four reference summaries. These SCUs are used to create tasks for the system evaluation phase.

Recall that in the original Pyramid, SCUs are exhaustively collected; then, coreferring SCUs \emph{between} reference summaries are merged and weighted by the number of reference summaries from which they originate.
In contrast, our method enables using a \textit{sample} of SCUs for evaluation, out of the SCUs collected in this phase (we have sampled, for uniformity, 32 SCUs per topic). Further, it avoids posing the task of merging coreferring SCUs \textit{across} references, which is difficult and error-prone, particularly when expected from crowd workers. Instead, we rely on the higher likelihood of a repeated fact to be included in our sample, possibly more than once. This implicitly increases the expected impact of repeated facts on our evaluation.

\paragraph*{System evaluation.}
In this phase, a crowd worker is presented with a system summary and a fixed-sized small set of SCUs (we used sets of~16 SCUs). The worker is asked whether each SCU can be inferred from the system summary text. The guidelines advise workers to refrain from using general knowledge and to ignore minor content differences between the SCU and the system summary.
Each SCU should be assessed by a few crowd workers, to ensure the stability of the results (in our experiments, each SCU was assigned for evaluation to~5 workers).

\paragraph*{Scoring.}
Following common practice in crowdsourcing, we use techniques of filtering out noisy workers who had high disagreement with others (pairwise worker agreement $<$ 0.5).
Then, using the remaining answers, we take the majority vote for each SCU to decide whether it appears in the system summary.\footnote{In our experiments, we have also examined the option of using the average answer, which was significantly worse.} We resolve ties with a ``not present'' default, as the more likely answer.
We then compute the system summary score as the percentage of SCUs it matched out of the set of judged SCUs. A system's final score is its average score over all topics.
\label{sec_method}

\section{Experiments}
\paragraph*{Experimental setup.}
We used the DUC 2005 and 2006 multi-document summarization datasets \citep{Nist2014Duc}, which contain expert evaluations for both Pyramid and Responsiveness. 
Each of the two datasets includes 20 document clusters, each pertaining to a target topic, with four reference summaries and 25 (2005) or 22 (2006) system summaries per topic. All summaries are 250 words long. On average, 105 weighted SCUs were extracted, by experts, for each topic. In comparison, our setup gathers~32 sampled crowdsourced unweighted SCUs.

As suggested in \citet{dang2006overviewOfDuc2006} and \citet{passonneau2006pyramidsInDuc}, the 2005 data tends to be easier to evaluate than the 2006 data, seemingly due to ``less natural" document clusters with respect to practical summarization settings.
\citet{passonneau2006pyramidsInDuc} show that the document sets in 2005 were overall more difficult for systems to summarize, as reflected by a lower average Pyramid score across all systems. The 2005 topics are more complex as they yield fewer general, context-independent SCUs. For example, as \citet{dang2006overviewOfDuc2006} indicates, there are more topics that had a relatively large number of specific named entities.
Consequently, due to the topic hardness, \citet{passonneau2006pyramidsInDuc} indicate very few significant differences between overall system Pyramid scores, as evident by Tukey's HSD test. While 2006 systems can be divided into eight significantly different Pyramid score groups, in 2005 only two such groups emanate.
Additionally, the guidelines and scoring method were slightly improved in 2006, relative to 2005.
For these reasons, we focused on the 2006 dataset, fully annotating it, while utilizing half the topics, randomly chosen, from the 2005 data.

Using Amazon Mechanical Turk,\footnote{\url{https://www.mturk.com/}} we qualified workers with over 5000 approved assignments and a 99\% approval rate. We paid workers \$0.50 per reference summary annotation assignment (generating 8 SCUs), yielding a total Pyramid creation cost of \$48 (including fees) for the 2005 dataset (10 topics) and \$96 for 2006 (20 topics). Pyramid creation cost per topic is thus \$4.8.
For the system summary evaluation phase we split the~32 SCUs to two tasks of~16 SCUs each, in order to ensure that the crowdsourcing platform assigns each SCU to~5 distinct workers. We paid workers \$0.45, and evaluated all 25 (2005) and 22 (2006) systems.
The total benchmark evaluation cost was \$1350 (including fees) for 2005 and \$2376 for 2006, equaling \$5.4 per system per topic, or \$108 per system evaluation over all 20 topics.

We release\footnote{\url{https://github.com/OriShapira/LitePyramids}} our SCU dataset for DUC 2005 and DUC 2006 as a complementary resource, accompanied by the HTML pages for our tasks on Amazon Mechanical Turk and processing and evaluation scripts. In the SCU dataset, we mark the SCUs we used in our experiments, including their grouping as tasks in the system evaluation phase. These enable future crowdsourced Pyramid evaluations of new systems on these datasets, as well as developing new datasets with crowdsourced pyramids.

\paragraph*{Correlations with original Pyramid.}
We first assess our evaluation methodology by computing the correlation of its system scores (and rankings) to those of the original Pyramid. These are compared with the analogous correlations for the expert Responsiveness scores, available in the datasets.
As seen in Table \ref{table_correlations}, our method produces better correlations, and substantially so on the more characteristic 2006 dataset.
Importantly, notice that Responsiveness scores here were obtained by \textit{experts}, and therefore the gap for crowdsourced Responsiveness is expected to be greater, further indicating the advantage of our method as a crowdsourcable approach.

\begin{table}[t!]
\centering
\resizebox{0.9\columnwidth}{!}{

\begin{tabular}{@{}c||c|c||c|c@{}} %

& \multicolumn{2}{c||}{Pearson ($\rho_p$)} & \multicolumn{2}{c}{Spearman ($\rho_s$)} \\\cline{2-5}
& Ours &  Expert Resp. & Ours & Expert Resp. \\ \hline
2005 & 0.81 & 0.81 & \textbf{0.79} & 0.77 \\ \hline
2006 & \textbf{0.74} & 0.60 & \textbf{0.69} & 0.40 \\ 

\end{tabular}}

\caption{Correlations to the original Pyramid scores, for our crowdsourced method and for \textit{expert} Responsiveness method, for DUC '05 and '06.}
\label{table_correlations}
\end{table}

\paragraph*{Stability.} As an additional assessment, we test the robustness of our method, in terms of its reproducibility. To that end, we reran the system evaluation phase on eight randomly chosen systems of the 2006 data, which enabled us to compare our results with those obtained by~\citet{gillick2010crowdsourcing} for crowdsourced Responsiveness for a similar setting (8 random systems of the 2006 dataset).
Notably, the lightweight Pyramid obtained an average 10\% relative change in overall system scores, whereas crowdsourced Responsiveness exhibited lower stability with an average of 24\% relative change.

\paragraph*{Cost analysis.}
We analyze the impact of randomly reducing the various resources involved in our methodology, aiming to see whether overall cost might be reduced without harming correlation with the original Pyramid. The results below, reported as averages over 70 re-sampled iterations for each setting, suggest that such cost reductions would be harmful. 

\textbf{Number of workers.}~~Reducing the number of workers per SCU judgment from five to three drops the correlations by about 8 points in 2006 and 6 points in 2005.

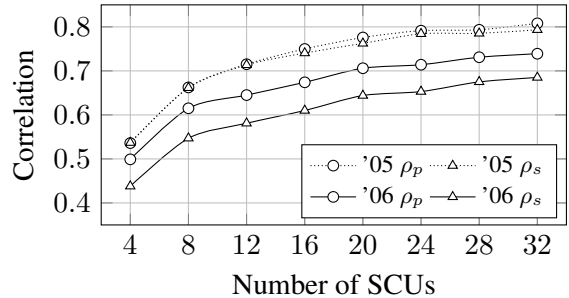
\begin{figure}[!t]

\begin{tikzpicture}

    \begin{axis}[
    axis on top,
    width=\linewidth,
        grid,
      legend cell align=left,
      legend columns=2,
      legend style={at={(0.98,0.03)},anchor=south east,font=\small},
        height=4.5cm,
        xlabel=Number of SCUs,
        ylabel=Correlation,
       xmin=2,   xmax=34,
    ymin=0.35,   ymax=0.85,
        xtick={4,8,...,32},
    ytick={0,0.1,0.2,...,1.0},
    ]

    \addplot[   densely dotted,
                color=black,
                mark=*,
                mark options={solid, fill=white},
                smooth] coordinates {
        (4, 0.535977767)
        (8, 0.662421239)
        (12, 0.715109566)
        (16, 0.749719389)
        (20, 0.775823481)
        (24, 0.791331494)
        (28, 0.793317079)
        (32, 0.808020524)
    };
    
    \addplot[   densely dotted,
                color=black,
                mark=triangle*,
                mark options={solid, fill=white},
                smooth] coordinates {
        (4, 0.536922584)
        (8, 0.661568623)
        (12, 0.713635811)
        (16, 0.74051723)
        (20, 0.762468363)
        (24, 0.784259726)
        (28, 0.785230699)
        (32, 0.793455613)
    };

    \addplot[   color=black,
                mark=*,
                mark options={solid, fill=white},
                smooth] coordinates {
        (4,0.499)
        (8,0.615)
        (12,0.645)
        (16,0.674)
        (20,0.706)
        (24,0.714)
        (28,0.731)
        (32,0.739)
    };
    
    \addplot[   color=black,
                mark=triangle*,
                mark options={solid, fill=white},
                smooth] coordinates {
        (4,0.438)
        (8,0.547)
        (12,0.581)
        (16,0.610)
        (20,0.644)
        (24,0.653)
        (28,0.675)
        (32,0.685)
    };

    \legend{'05 $\rho_p$,'05 $\rho_s$,'06 $\rho_p$,'06 $\rho_s$}

\end{axis}

\end{tikzpicture}

\caption{Average Pearson and Spearman correlations with Pyramid scores as a function of number of SCUs evaluated per topic, on the DUC '05 and '06 data. }
\label{fig_graphCorrByScuCount}

\end{figure}

\textbf{Number of SCUs.}~~Figure \ref{fig_graphCorrByScuCount} shows that correlation increases as a function of the number of judged SCUs per topic. The correlation improvement seems to stabilize around 32 SCUs.

\begin{figure}[!t]

\begin{tikzpicture}

    \begin{axis}[
    axis on top,
    width=\linewidth,
        grid,
      legend cell align=left,
      legend columns=2,
      legend style={at={(0.98,0.03)},anchor=south east,font=\small},
        height=4.5cm,
        xlabel=Number of Topics,
        ylabel=Correlation,
       xmin=1,   xmax=21,
    ymin=0.35,   ymax=0.85,
        xtick={2,4,...,20},
    ytick={0,0.1,0.2,...,1.0},
    ]
    
    \addplot[   densely dotted,
                color=black,
                mark=*,
                mark options={solid, fill=white},
                smooth] coordinates {
        (2, 0.680503072)
        (4, 0.748044387)
        (6, 0.772848733)
        (8, 0.799227306)
        (10, 0.808020524)
    };
    
    \addplot[   densely dotted,
                color=black,
                mark=triangle*,
                mark options={solid, fill=white},
                smooth] coordinates {
        (2, 0.663269247)
        (4, 0.716357518)
        (6, 0.750850318)
        (8, 0.770063454)
        (10, 0.793455613)
    };

    \addplot[   color=black,
                mark=*,
                mark options={solid, fill=white},
                smooth] coordinates {
        (2, 0.461713589)
        (4, 0.520127561)
        (6, 0.565458478)
        (8, 0.595533543)
        (10, 0.652952523)
        (12, 0.670361295)
        (14, 0.685196674)
        (16, 0.71799842)
        (18, 0.723820715)
        (20, 0.739329373)
    };
    
    \addplot[   color=black,
                mark=triangle*,
                mark options={solid, fill=white},
                smooth] coordinates {
        (2, 0.41661099)
        (4, 0.479874704)
        (6, 0.510289249)
        (8, 0.516950481)
        (10, 0.55640229)
        (12, 0.592439638)
        (14, 0.606162238)
        (16, 0.633829374)
        (18, 0.650421696)
        (20, 0.68456799)
    };

    \legend{'05 $\rho_p$,'05 $\rho_s$,'06 $\rho_p$,'06 $\rho_s$}

\end{axis}

\end{tikzpicture}

\caption{Average Pearson and Spearman correlations with Pyramid scores as a function of number of topics used for evaluation, on the DUC '05 and '06 data.}
\label{fig_graphCorrByEventCount}

\end{figure}
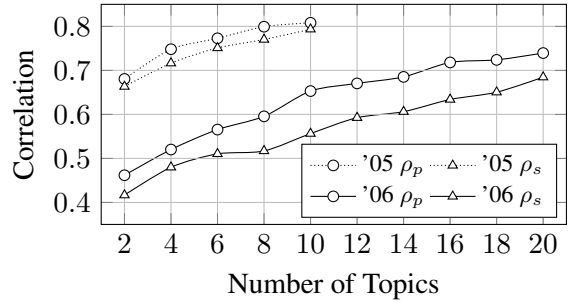

\textbf{Number of topics.}~~Figure \ref{fig_graphCorrByEventCount} presents the effect of the number of topics on which systems are evaluated, showing a steady correlation increase, which does not necessarily saturate at the number of 20 topics available in these datasets.


\paragraph*{Qualitative analysis.}
To identify certain limitations of our methodology, we manually analyzed some ``suspected" topics, for which either worker Krippendorff agreement or correlation with the original Pyramid was low. We noticed two interesting phenomena.

First, some topics seem inherently more difficult to evaluate, particularly for crowd workers. 
Such difficulty may be attributed to SCUs that are more difficult to assess or to less coherent system summaries, due to the respective document set's complexity.
Indeed, \citet{passonneau2006pyramidsInDuc} indicated that topic characteristics and annotator training experience effect evaluation quality.
It seems worthwhile investigating, in future research, whether correlations improve
by increasing further the overall number of topics, reducing the impact of the problematic ones.

Another possibility may be to filter out topics with low annotator agreement when computing systems' scores by the lightweight Pyramid method. We hypothesize that doing so might improve the reliability of this method, and hence increase its correlation with the original, expert-based, Pyramid method (when the latter is computed over all test topics).
Indeed, in a preliminary test, we filtered out those 20\% of the topics with lowest Krippendorff annotator agreement. This yielded a 6-point Spearman score increase (relative to the correlations reported in Table \ref{table_correlations}) when correlated with the original Pyramid ranking, as computed over the full set of topics. We note that while Figure \ref{fig_graphCorrByEventCount} shows a slight decrease in average correlation when removing 4 random topics, removing specifically the 4 low-agreement topics seems to improves it notably. Further analysis might conclude that filtering problematic topics generically improves the reliability of the lightweight Pyramid method.

The second phenomenon observed among the difficult topics was that in some, the 32 sampled SCUs seem to miss important information, causing an unjustified degradation in system scores. In analogy to the variance in the number of SCUs in exhaustive Pyramids, it would be interesting to investigate methods for varying the sample size in our lightweight approach, based on some automatically detected parameters of topic complexity.
\label{sec_results}

\section{Conclusion and Future Work}
To the best of our knowledge, our method is the first to mimic the reliable Pyramid method as an affordable crowdsourced procedure. Our experiments suggest that this lightweight Pyramid is more reliable than the common Responsiveness method. It also allows comparing multiple systems with absolute scores, which pairwise comparison does not.

Future work may improve correlation with the original Pyramid, or reduce annotation cost, by following our qualitative analysis and by reducing crowdsourcing noise (via  qualification tests, enhanced guidelines, and post-processing result normalization  \citep{hovy2013mace, plank2014crowdsourcing, hosseini2012crowdsourcing}).
It would be appealing to investigate applying our methods to additional evaluation datasets, for which original Pyramid evaluations are not available for comparison. For example, addressing the CNN/DailyMail dataset \citep{Nallapati2016CnnDM} would involve testing single document summarization,  utilizing a single reference summary per source text and addressing varying lengths of reference and system summaries.

The Pyramid method is mainly a measurement of recall, which thus also applies to our lightweight Pyramid; but other measurements for summary quality, such as precision, non-redundancy and grammaticality, may also be considered. In particular, it may be possible to extend our design of crowdsourcing tasks to supply indications for these complementary measurements as well.
\label{sec_conclusion}

\section*{Acknowledgments}

We would like to thank the anonymous reviewers for their constructive comments, as well as Ani Nenkova for her helpful remarks.
This work was supported in part by the Bloomberg Data Science Research Grant Program; by the German Research Foundation through the German-Israeli Project Cooperation (DIP, grants DA 1600/1-1 and GU 798/17-1); by the BIU Center for Research in Applied Cryptography and Cyber Security in conjunction with the Israel National Cyber Bureau in the Prime Minister's Office; by the Israel Science Foundation (grants 1157/16 and 1951/17); by DARPA Young Faculty Award YFA17-D17AP00022; and by the ArguAna Project GU 798/20-1 (DFG).

\bibliography{main}

\begin{thebibliography}{22}
\expandafter\ifx\csname natexlab\endcsname\relax\def\natexlab#1{#1}\fi

\bibitem[{Chaganty et~al.(2018)Chaganty, Mussmann, and
  Liang}]{chaganty2018price}
Arun Chaganty, Stephen Mussmann, and Percy Liang. 2018.
\newblock The price of debiasing automatic metrics in natural language
  evalaution.
\newblock In \emph{Proceedings of the 56th Annual Meeting of the Association
  for Computational Linguistics (Volume 1: Long Papers)}, volume~1, pages
  643--653.

\bibitem[{Dang(2006)}]{dang2006overviewOfDuc2006}
Hoa~Trang Dang. 2006.
\newblock Overview of duc 2006.
\newblock In \emph{Proceedings of the document understanding conference}.

\bibitem[{Falke et~al.(2017)Falke, Meyer, and
  Gurevych}]{Falke2017ConceptMapMDS}
Tobias Falke, Christian~M. Meyer, and Iryna Gurevych. 2017.
\newblock Concept-map-based multi-document summarization using concept
  coreference resolution and global importance optimization.
\newblock In \emph{Proceedings of the 8th International Joint Conference on
  Natural Language Processing}, pages 801--811.

\bibitem[{Fan et~al.(2018)Fan, Grangier, and Auli}]{Fan2018ControllableAS}
Angela Fan, David Grangier, and Michael Auli. 2018.
\newblock Controllable abstractive summarization.
\newblock In \emph{Proceedings of the 2nd Workshop on Neural Machine
  Translation and Generation}, pages 45--54. Association for Computational
  Linguistics.

\bibitem[{Gao et~al.(2018)Gao, Meyer, and Gurevych}]{Gao2018Preference}
Yang Gao, Christian~M. Meyer, and Iryna Gurevych. 2018.
\newblock April: Interactively learning to summarise by combining active
  preference learning and reinforcement learning.
\newblock In \emph{Proceedings of the 2018 Conference on Empirical Methods in
  Natural Language Processing}, pages 4120--4130. Association for Computational
  Linguistics.

\bibitem[{Gillick and Liu(2010)}]{gillick2010crowdsourcing}
Dan Gillick and Yang Liu. 2010.
\newblock Non-expert evaluation of summarization systems is risky.
\newblock In \emph{Proceedings of the NAACL HLT 2010 Workshop on Creating
  Speech and Language Data with Amazon's Mechanical Turk}, pages 148--151.
  Association for Computational Linguistics.

\bibitem[{Grusky et~al.(2018)Grusky, Naaman, and Artzi}]{Grusky2018Newsroom}
Max Grusky, Mor Naaman, and Yoav Artzi. 2018.
\newblock Newsroom: A dataset of 1.3 million summaries with diverse extractive
  strategies.
\newblock In \emph{Proceedings of the 2018 Conference of the North American
  Chapter of the Association for Computational Linguistics: Human Language
  Technologies}, pages 708--719, New Orleans, Louisiana. Association for
  Computational Linguistics.

\bibitem[{Hirao et~al.(2018)Hirao, Kamigaito, and
  Nagata}]{hirao2018autoPyramids}
Tsutomu Hirao, Hidetaka Kamigaito, and Masaaki Nagata. 2018.
\newblock Automatic pyramid evaluation exploiting edu-based extractive
  reference summaries.
\newblock In \emph{Proceedings of the 2018 Conference on Empirical Methods in
  Natural Language Processing}, pages 4177--4186.

\bibitem[{Hosseini et~al.(2012)Hosseini, Cox, Mili{\'c}-Frayling, Kazai, and
  Vinay}]{hosseini2012crowdsourcing}
Mehdi Hosseini, Ingemar~J Cox, Nata{\v{s}}a Mili{\'c}-Frayling, Gabriella
  Kazai, and Vishwa Vinay. 2012.
\newblock On aggregating labels from multiple crowd workers to infer relevance
  of documents.
\newblock In \emph{European Conference on Information Retrieval}, pages
  182--194. Springer.

\bibitem[{Hovy et~al.(2013)Hovy, Berg-Kirkpatrick, Vaswani, and
  Hovy}]{hovy2013mace}
Dirk Hovy, Taylor Berg-Kirkpatrick, Ashish Vaswani, and Eduard Hovy. 2013.
\newblock Learning whom to trust with mace.
\newblock In \emph{Proceedings of the 2013 Conference of the North American
  Chapter of the Association for Computational Linguistics: Human Language
  Technologies}, pages 1120--1130.

\bibitem[{Lin(2004)}]{lin2004rouge}
Chin-Yew Lin. 2004.
\newblock Rouge: A package for automatic evaluation of summaries.
\newblock \emph{Text Summarization Branches Out}.

\bibitem[{Nallapati et~al.(2016)Nallapati, Zhou, dos Santos, Gulcehre, and
  Xiang}]{Nallapati2016CnnDM}
Ramesh Nallapati, Bowen Zhou, Cicero dos Santos, Caglar Gulcehre, and Bing
  Xiang. 2016.
\newblock Abstractive text summarization using sequence-to-sequence rnns and
  beyond.
\newblock In \emph{Proceedings of The 20th SIGNLL Conference on Computational
  Natural Language Learning}, pages 280--290. Association for Computational
  Linguistics.

\bibitem[{Nenkova and Passonneau(2004)}]{nenkova2004pyramids}
Ani Nenkova and Rebecca Passonneau. 2004.
\newblock Evaluating content selection in summarization: The pyramid method.
\newblock In \emph{Proceedings of the human language technology conference of
  the north american chapter of the association for computational linguistics:
  Hlt-naacl 2004}.

\bibitem[{{NIST}(2003)}]{Nist2003Duc2003Tasks}
{NIST}. 2003.
\newblock Duc 2003: Documents, tasks, and measures.
\newblock https://duc.nist.gov/duc2003/tasks.html.

\bibitem[{{NIST}(2014)}]{Nist2014Duc}
{NIST}. 2014.
\newblock Document understanding conferences.
\newblock https://duc.nist.gov/.

\bibitem[{{NIST}(2018)}]{Nist2018Tac}
{NIST}. 2018.
\newblock Text analysis conference.
\newblock https://tac.nist.gov/.

\bibitem[{Owczarzak et~al.(2012)Owczarzak, Conroy, Dang, and
  Nenkova}]{owczarzak2012autoEvalAssessment}
Karolina Owczarzak, John~M Conroy, Hoa~Trang Dang, and Ani Nenkova. 2012.
\newblock An assessment of the accuracy of automatic evaluation in
  summarization.
\newblock In \emph{Proceedings of Workshop on Evaluation Metrics and System
  Comparison for Automatic Summarization}, pages 1--9. Association for
  Computational Linguistics.

\bibitem[{Passonneau(2006)}]{Passonneau2006PyramidGuide}
Rebecca Passonneau. 2006.
\newblock Pyramid annotation guide: Duc 2006.
\newblock
  http://www1.cs.columbia.edu/~becky/DUC2006/2006-pyramid-guidelines.html.

\bibitem[{Passonneau et~al.(2006)Passonneau, McKeown, Sigelman, and
  Goodkind}]{passonneau2006pyramidsInDuc}
Rebecca Passonneau, Kathleen McKeown, Sergey Sigelman, and Adam Goodkind. 2006.
\newblock Applying the pyramid method in the 2006 document understanding
  conference.

\bibitem[{Paulus et~al.(2018)Paulus, Xiong, and Socher}]{paulus2017deep}
Romain Paulus, Caiming Xiong, and Richard Socher. 2018.
\newblock A deep reinforced model for abstractive summarization.
\newblock \emph{Sixth International Conference on Learning Representations}.

\bibitem[{Plank et~al.(2014)Plank, Hovy, and
  S{\o}gaard}]{plank2014crowdsourcing}
Barbara Plank, Dirk Hovy, and Anders S{\o}gaard. 2014.
\newblock Linguistically debatable or just plain wrong?
\newblock In \emph{Proceedings of the 52nd Annual Meeting of the Association
  for Computational Linguistics (volume 2: Short Papers)}, volume~2, pages
  507--511.

\bibitem[{Yang et~al.(2016)Yang, Passonneau, and de~Melo}]{yang2016PEAK}
Qian Yang, Rebecca~J. Passonneau, and Gerard de~Melo. 2016.
\newblock Peak: Pyramid evaluation via automated knowledge extraction.
\newblock In \emph{Proceedings of the 30th {AAAI} Conference on Artificial
  Intelligence (AAAI 2016)}. {AAAI} Press.

\end{thebibliography}
\bibliographystyle{acl_natbib}

\end{document}